# A Method for Determining Weights of Criterias and Alternative of Fuzzy Group Decision Making Problem

Jon JaeGyong, Mun JongHui, Ryang GyongIl

Faculty of Electronics & Automation, **Kim Il Sung** University,

Pyongyang, Democratic People's Republic of Korea

**Abstract:** In this paper, we constructed a model to determine weights of criterias and presented a solution for determining the optimal alternative by using the constructed model and relationship analysis between criterias in fuzzy group decision-making problem with different forms of preference information of decision makers on criterias.

**1. Introduction**

Determining optimal alternative in group decision making is very important in management.

Up to now, there are many researches for multipleobjective decision making problems that have more than one optimize purpose, fuzzy decision making problem in which information of alternative on criteria are fuzzy number, group decision making problem that have many decision makers.

To be concrete as follows:

Some researchers[1,2] paid their attention to group decision-making problems, in which decision-makers provide their preference information on alternative using different preference forms.

Literature[12] further established a single-objective optimization model to determine a optimal alternative from interval decision matrix and criteria weight intervals.

There are methods that get the optimal alternative by synthesizing conclusions of decision-makers when information of criteria weight is given partially and degree of satisfaction on criterias are expressed by triangle fuzzy number[3,4,5].

Literature[13] proposed a method for determining the preference of alternative by Borda's method based on preference of interval number when information of decision makers on evaluation alternative is stated by interval number.

Literature[11] proposed a new method for solution of Fuzzy Multiple Criteria Decision Making by introducing score function G that remove the defect of S function and H function.

In general, there are benefit attributes and cost attributes in multiple-attribute decision-making problems, and dimention of different attributes may be different.

Literatures[8,9] proposed a method that determine the optimal attribute based on Satisfactory Matrix $\Pi = (\mu_{ij})_{m \times n}$ which is mapped from D according to criteria type(benefit attribute or cost attribute)

Literatures[6,7] present a group decision making method that decreases disagreement of decision-makers by fuzzy programming method when required information is given as a supplementary decision matrix.

Literature[10] divided decision making problem into several subproblems when composition of decision makers is very complex and scale of it is very large, and presented a method that determines the solution of decision making problem based on the solution of every subproblem.

The shortcomings of preceding methods for fuzzy multi-criteria group decision-making are as follows:

First, in fuzzy multi-criteria group decision making problem, a method to determine of criterias' weight was supported in case that required information of decision makers were given in different forms, but if any decision makers gave their required information in form of multiplicative preference relation, then supported model can not include in full requirement of decision makers.

Second, a method to determine optimal alternative based on relationship analysis between criterias was supported in case that satisfactory matrixes were given by triangle fuzzy number in fuzzy multi-criteria group decision making problem. But dividing algorithm of criterias that is subalgorithm of optimal alternative determining algorithm unilaterally divides two subsets of all criterias, so conflict or irrelevant criterias may be in one subset.

Third, in case that satisfactory matrix is given by triangle fuzzy number, final degree for evaluation of alternatives is also triangle fuzzy number, so, satisfactory degree of alternative on criterias is not quantitativelly expressed.

For this reason, in the second part of this paper, we propose a model to determine weights reasonably in consideration of left or right limited value in case that there are decision makers who give their required information in form of multiplicative preference relation in fuzzy multi-criterias group decision making problem.

In the third part we newly support a dividing algorithm of criterias' set for determing optimal alternative, thus all criterias of one subset cooperate with each other, and then a final evaluation degree to evaluate alternatives based on relationship analysis between criterias in case that satisfactory matrixes are represented by triangle fuzzy number.

In the fourth part we verified the effectiveness of supported methods by solving example.

## 2. A METHOD FOR DETERMINING WEIGHTS OF CRITERIAS

### 2.1 Type of Information Expression of Decision-makers on criteria weights

In group decision-making problem, a decision-maker can represent weight information of criteria as one of the following three different expression forms[9].

**First**, Utility values $\tilde{U} = \{\tilde{u}_1, \tilde{u}_2 \cdots \tilde{u}_n\}$: $\tilde{u}_j = (ul_j, um_j, uu_j)$ indicates utility interval of criteria weights, where $0 \le ul_j \le um_j \le uu_j \le 1$ ($j = \overline{1,n}$), and $ul_j, um_j, uu_j$ are infimum, median, supremum of weight of criteria $c_j$.

**Second,** Preference relation $\tilde{P} = (\tilde{p}_{ij})_{n \times n}$: $\tilde{p}_{ij} = (pl_{ij}, pm_{ij}, pu_{ij})$ indicates the preference degree of criteria $c_i$ over $c_j$ and is the preference information on $p_{ij}$ that is described by the following formula

$$p_{ij} = 0.5(w_i - w_j + 1),$$

Where $pl_{ij}$, $pm_{ij}$, $pu_{ij}$ are infimum, median, supremum of $p_{ij}$, and $w_i, w_j$ are weights of criteria $c_i$, $c_j$.

From the definition, the following condition is satisfied

$$0 \leq pl_{ij} \leq pm_{ij} \leq pu_{ij} \leq 1, \quad pl_{ij} + pu_{ji} = pm_{ij} + pm_{ji} = pu_{ij} + pl_{ji} = 1, \quad pl_{ii} = pm_{ii} = pu_{ii} = 0.5 \quad (i,j = \overline{1,n})$$

**Third**, Multiplicative preference relation $\tilde{B} = (\tilde{b}_{ij})_{n \times n}$ : $\tilde{b}_{ij} = (bl_{ij}, bm_{ij}, bu_{ij})$ indicates multiplicative preference degree of criteria $c_i$ over $c_j$ and is the preference information on $b_{ij}$ that is described by the following formula

$$b_{ij} = \frac{w_i}{w_j}$$

where $bl_{ij}, bm_{ij}, bu_{ij}$ are infimum, median, supremum of $b_{ij}$, and $w_i$, $w_j$ are weights of criteria $c_i$, $c_j$.

From the definition, the following condition is satisfied.

$$0 < bl_{ij} \leq bm_{ij} \leq bu_{ij}, \quad bl_{ij}bu_{ji} = bm_{ij}bm_{ji} = bu_{ij}bl_{ji} = 1, \quad bl_{ii} = bm_{ii} = bu_{ii} = 1 \quad (i,j = \overline{1,n})$$

Above three expression forms are the most general forms that can represent criteria weight information.

### 2.2 A Method For Determining Weights of Criterias

In decision-making problem, let $w_1, w_2, \cdots, w_n$ be the determined weight of criterias.

If these weights satisfy the requirement of all decision-makers that are represented above three forms, the following condition must be satisfied for $i, j = \overline{1,n}$

$$ul_j \leq w_j \leq uu_j \tag{1}$$

$$(w_j - um_j)^2 = em_j \leq \varepsilon_{j1} \tag{2}$$

$$pl_{ij} \leq 0.5(w_i - w_j + 1) \leq pu_{ij} \tag{3}$$

$$\left(0.5(w_i - w_j + 1) - pm_{ij}\right)^2 = em_{ij} \leq \varepsilon_{ij2} \tag{4}$$

$$bl_{ij} \leq \frac{w_i}{w_j} \leq bu_{ij} \tag{5}$$

$$\left(\frac{w_i}{w_j} - bm_{ij}\right)^2 = em_{ij} \leq \varepsilon_{ij3} \quad (w_j \neq 0) \tag{6}$$

where $\varepsilon_{j1}$, $\varepsilon_{ij2}$, $\varepsilon_{ij3}$ are nonnegative real numbers that is small sufficiently.

Generally, it is difficult to satisfy all requirements of decision-makers.

Therefore, we must determine the reasonable criteria weight that relax the contradiction of information as possible and satisfy requirements of the decision-makers as possible.

For group decision-making problems with t decision-makers, without loss of generality, we suppose that decision maker $k$ ($k=1,\ldots, t_1$) provides his preference information by means of the utility values, decision maker $k$ ($k=t_1+1,\ldots, t_2$) provides his preference information by means of the preference relation and decision maker $k$ ($k=t_2+1,\ldots, t$) provides his preference information by means of the multiplicative preference relation.

In order to determine the weight vector of criterias, we establish the following linear programming problem.

$$[M] \quad E = \min\left(\sum_{j=1}^{n}\sum_{d=1}^{t_1} w_{J_d}(el_j^{(d)} + em_j^{(d)} + eu_j^{(d)}) + \sum_{i,j=1}^{n}\sum_{d=t_1+1}^{t} w_{J_d}(el_{ij}^{(d)} + em_{ij}^{(d)} + eu_{ij}^{(d)})\right) \quad (7)$$

s.t. $em_j^{(d)} = (w_j - um_j^{(d)})^2,\ j=\overline{1,n},\ d=\overline{1,t_1}$

$w_j \geq ul_j^{(d)} - el_j^{(d)},\ j=\overline{1,n},\ d=\overline{1,t_1}$
$w_j \leq uu_j^{(d)} + eu_j^{(d)},\ j=\overline{1,n},\ d=\overline{1,t_1}$
$el_j^{(d)}, eu_j^{(d)} \geq 0,\ j=\overline{1,n},\ d=\overline{1,t_1}$
$em_{ij}^{(d)} = (0.5(w_i - w_j + 1) - pm_{ij}^{(d)})^2,\ i,j=\overline{1,n},\ d=\overline{t_1+1,t_2}$
$0.5(w_i - w_j + 1) \geq pl_{ij}^{(d)} - el_{ij}^{(d)},\ i,j=\overline{1,n},\ d=\overline{t_1+1,t_2}$
$0.5(w_i - w_j + 1) \leq pu_{ij}^{(d)} + eu_{ij}^{(d)},\ i,j=\overline{1,n},\ d=\overline{t_1+1,t_2}$
$em_{ij}^{(d)} = \left(\dfrac{w_i}{w_j} - bm_{ij}^{(d)}\right)^2,\ i,j=\overline{1,n},\ d=\overline{t_2+1,t}$
$\dfrac{w_i}{w_j} \geq bl_{ij}^{(d)} - el_{ij}^{(d)},\ i,j=\overline{1,n},\ d=\overline{t_2+1,t}$
$\dfrac{w_i}{w_j} \leq bu_{ij}^{(d)} + eu_{ij}^{(d)},\ i,j=\overline{1,n},\ d=\overline{t_2+1,t}$
$el_{ij}^{(d)}, eu_{ij}^{(d)} \geq 0,\ i,j=\overline{1,n},\ d=\overline{t_1+1,t}$
$\sum_{j=1}^{n} w_j = 1,\ w_j > 0,\ j=\overline{1,n}$

Where $el_j^{(d)}$, $eu_j^{(d)}$ ($j=\overline{1,n},\ d=\overline{1,t_1}$), $el_{ij}^{(d)}$, $eu_{ij}^{(d)}$ ($i,j=\overline{1,n},\ d=\overline{t_1+1,t}$) are preference information deviation value of decision maker $J_d$ and $w_{J_d}$ is weight of decision maker $J_d$

Decision makers can adjust the preference information with nonzero deviation values by compromising their requirement. Adjusting method is that it give a limited value at first, and then turn up supremum a little, turn down infimum a little on equal terms with it doesn't exceed limited interval. At this time amount of change is proportionate with deviation value and extends the range of triangle fuzzy number that represents required information a little wide.

And we establish the model again on the basis of changed preference information value and get the deviation values again.

If this procedure is repeated until all of the deviation values come to zero, the final weight values is just the optimal criteria weight vector satisfying the requirement of every decision makers.

### 3. A METHOD FOR DETERMINING THE OPTIMAL ALTERNATIVE

For Fuzzy Multi-Criteria Group Decision Making Problem, here consider a new methodology for selecting the appropriate alternative among feasible alternative set A in the case that $d_i^j$ ($i=\overline{1,m},\ j=\overline{1,n}$) is given by triangular fuzzy number, decision matrix D is mapping to the matrix of degree of satisfaction and weight of decision makers is given as follows:

$$W_J = (w_{J_1}, w_{J_2}, \cdots, w_{J_t}), \qquad \sum_{d=1}^{t} w_{J_d} = 1$$

The the matrix of degree of satisfaction of decision maker $J_d$ ($d = \overline{1,t}$) is expressed as follows:

$$\tilde{\Pi}^{(d)} = \left(\mu_{ij}^{(d)}(x)\right)_{m \times n} = \begin{pmatrix} \mu_{11}^{(d)}(x) & \mu_{12}^{(d)}(x) & \cdots & \mu_{1n}^{(d)}(x) \\ \mu_{21}^{(d)}(x) & \mu_{22}^{(d)}(x) & \cdots & \mu_{2n}^{(d)}(x) \\ & \cdots & & \\ \mu_{m1}^{(d)}(x) & \mu_{m2}^{(d)}(x) & \cdots & \mu_{mn}^{(d)}(x) \end{pmatrix}, (d = \overline{1,t}) \qquad (8)$$

where $\mu_{ij}^{(d)}(x) = (\alpha_{ij}^{(d)}, \beta_{ij}^{(d)}, \gamma_{ij}^{(d)})$.

When there are several criterias for evaluating the alternatives, generally, some relationships between criterias exist.

If an increase of the degree of satisfaction for one criteria causes to increase the degree of satisfaction for other criteria, then a pair of two criterias are said to be cooperative. If an increase of the degree of satisfaction for one criteria causes to decrease the degree of satisfaction for other criteria, then a pair of two criterias are said to be conflictive. If an increase of the degree of satisfaction for one criterion doesn't cause the variation of the degree of satisfaction, then a pair of two criterias are said to be independent

**[Definition 1:]**

Define a triangular fuzzy number $\mu_{ijk}(x)$ as follows.

$$\mu_{ijk}(x) \stackrel{\Delta}{=} \mu_{ik}(x) \ominus \mu_{jk}(x)$$

where $\ominus$ denotes to extended subtraction of fuzzy numbers and $\mu_{ijk}(x)$ indicates the degree of satisfaction of alternative $a_i$, $a_j$ for criteria $c_k$.

Assume that $\mu_{ijk}(x)$ is a triangular fuzzy number $\mu_{ijk}(x) = (\alpha_{ijk}, \beta_{ijk}, \gamma_{ijk})$.

**[Difinition 2:]** (Cooperative degree and conflicting degree)

Cooperative degree: $\qquad cp(c_k, c_l) \stackrel{\Delta}{=} \dfrac{\sum_{(a_i, a_j) \in CP}(|\beta_{ijk}| + |\beta_{ijl}|)}{\sum_{(a_i, a_j) \in AP}(|\beta_{ijk}| + |\beta_{ijl}|)}$  (9)

Conflicting degree: $\qquad cf(c_k, c_l) \stackrel{\Delta}{=} \dfrac{\sum_{(a_i, a_j) \in CF}(|\beta_{ijk}| + |\beta_{ijl}|)}{\sum_{(a_i, a_j) \in AP}(|\beta_{ijk}| + |\beta_{ijl}|)}$  (10)

$c_k, c_l$: two criterias, $AP = \{(a_i, a_j) \mid \forall a_i, a_j \in A, i \neq j\}$: a set of alternatives pairs,

$$CP = \{(a_i, a_j) \mid \left[\int_{x>0} \mu_{ijk}(x) - \int_{x<0} \mu_{ijk}(x)\right] \times \left[\int_{x>0} \mu_{ijl}(x) - \int_{x<0} \mu_{ijl}(x)\right] > 0\}$$

$$CF = \{(a_i, a_j) \mid \left[\int_{x>0} \mu_{ijk}(x) - \int_{x<0} \mu_{ijk}(x)\right] \times \left[\int_{x>0} \mu_{ijl}(x) - \int_{x<0} \mu_{ijl}(x)\right] < 0\}$$

$$AP = CP \cup CF \cup IR$$

$$IR = \{(a_i, a_j) \mid \left[ \int_{x>0} \mu_{ijk}(x) - \int_{x<0} \mu_{ijk}(x) \right] \times \left[ \int_{x>0} \mu_{ijl}(x) - \int_{x<0} \mu_{ijl}(x) \right] = 0\}$$

$$CP \cap CF = CF \cap IR = CP \cap IR = \phi$$

In order to evaluate several alternatives for multi-criteria, devide overall set of criterias into two subsets based on conflicting, cooperative and irrelevant relation such that all criterias in each subset cooperate with each other.

The algorithm for dividing a set of criterias is as follows.

**[Step 1]** Setting the initial value.

$$S = \{c_2, c_3, \cdots, c_n\}, \quad q = 1, \quad S_1 = \{c_1\}$$

**[Step 2]** Calculation $p_j$ for all $c_j \in S$

$$p_j = \sum_{c_i \in S_q} \left( cp(c_i, c_j) - cf(c_i, c_j) \right)$$

**[Step 3]** Let be $p_k = \max_j p_j$, $k = \arg\max_j p_j$

**[Step 4]** If $p_k > 0$, then

$$S_q = S_q + \{c_k\}, \quad S = S - \{c_k\}$$

and go to step 2,

or else print $S_q$ and set as follows

$$q = q + 1, \quad S_q = \{c_k\}, \quad S = S - \{c_k\}$$

**[Step 5]** If $S = \phi$, then print $S_q$ and exit the algorithm, or else go to step 2.

This algorithm, first assigns the criteria $c_1$ to $S_1$, then takes away the criteria which belongs to $S_2$ and has maximum cooperative degree with criterias of $S_1$ to that subset, so we get one subset in which all criterias cooperate with each other. Then, we apply above introduced method to remaining criterias' set. In result, all criterias are classified into several classes.

Next the final degree of satisfaction is calculated as follows:

$$\mu^{(d)}(a_k) \overset{\Delta}{=} \sum_{c_i \in S_q^{(d)}} w_i^{(d)} \mu_{ki}^{(d)}(x) \tag{11}$$

where q satisfies following relation.

$$\sum_{c_i \in S_q^{(d)}} w_i^{(d)} = \max \left\{ \sum_{c_i \in S_1^{(d)}} w_i^{(d)}, \sum_{c_i \in S_2^{(d)}} w_i^{(d)}, \cdots, \sum_{c_i \in S_z^{(d)}} w_i^{(d)} \right\}$$

Next, we get an intergrated final degree of satisfaction by intergrating the final degree of satisfaction of all decision-makers.

The intergrated final degree of satisfaction is calculated as follows:

$$\mu(a_k) \overset{\Delta}{=} \sum_{d=1}^{t} w_{J_d} \mu^{(d)}(a_k) \tag{12}$$

where $w_{J_d}$ is a weight of decision-maker $J_d$.

An alternative that has maximum intergrated final degree of satisfaction is a optimal alternative.

## 4. A NUMERICAL ANALYSIS

Suppose that a set of decision makers is $J = \{J_1, J_2, J_3, J_4, J_5\}$, a set of alternatives is $A = \{a_1, a_2, a_3, a_4, a_5\}$, a set of criterias is $C = \{c_1, c_2, c_3, c_4\}$.

Decision maker 1 gives his requirement information for criteria weight in form of utility, decision maker 2 and 3 give in form of preference relation, decision maker 4 and 5 give in form of multiplicative preference relation.

The weight of decision makers is as follows:

$$W_J = \{w_{J_1}, w_{J_2}, w_{J_3}, w_{J_4}, w_{J_5}\} = \{0.2, 0.25, 0.2, 0.2, 0.15\}$$

The requirement information for criteria weight and the satisfactory matrix of decision makers is as follows.

$$\tilde{U}^{(1)} = \{\tilde{u}_1^{(1)}, \tilde{u}_2^{(1)}, \tilde{u}_3^{(1)}, \tilde{u}_4^{(1)}\} =$$
$$= \{(0.15, 0.20, 0.30), (0.05, 0.15, 0.20), (0.30, 0.35, 0.40), (0.20, 0.30, 0.35)\}$$

$$\tilde{P}^{(2)} = (\tilde{p}_{ij}^{(2)})_{4\times 4} = \begin{pmatrix} (0.50, 0.50, 0.50) & (0.20, 0.30, 0.35) & (0.25, 0.30, 0.40) & (0.50, 0.55, 0.60) \\ (0.65, 0.70, 0.80) & (0.50, 0.50, 0.50) & (0.50, 0.65, 0.75) & (0.80, 0.85, 0.95) \\ (0.60, 0.70, 0.75) & (0.25, 0.35, 0.50) & (0.50, 0.50, 0.50) & (0.65, 0.70, 0.80) \\ (0.40, 0.45, 0.50) & (0.05, 0.15, 0.20) & (0.20, 0.30, 0.35) & (0.50, 0.50, 0.50) \end{pmatrix}$$

$$\tilde{P}^{(3)} = (\tilde{p}_{ij}^{(3)})_{4\times 4} = \begin{pmatrix} (0.50, 0.50, 0.50) & (0.25, 0.25, 0.30) & (0.30, 0.35, 0.40) & (0.15, 0.20, 0.25) \\ (0.70, 0.75, 0.75) & (0.50, 0.50, 0.50) & (0.55, 0.60, 0.65) & (0.40, 0.45, 0.50) \\ (0.60, 0.65, 0.70) & (0.35, 0.40, 0.45) & (0.50, 0.50, 0.50) & (0.25, 0.30, 0.40) \\ (0.75, 0.80, 0.85) & (0.50, 0.55, 0.60) & (0.60, 0.70, 0.75) & (0.50, 0.50, 0.50) \end{pmatrix}$$

$$\tilde{B}^{(4)} = (\tilde{b}_{ij}^{(4)})_{4\times 4} = \begin{pmatrix} (1.00, 1.00, 1.00) & (0.38, 0.40, 0.42) & (0.33, 0.34, 0.36) & (0.42, 0.43, 0.45) \\ (2.40, 2.50, 2.60) & (1.00, 1.00, 1.00) & (0.55, 0.59, 0.62) & (2.20, 2.30, 2.40) \\ (2.80, 2.90, 3.00) & (1.60, 1.70, 1.80) & (1.00, 1.00, 1.00) & (2.50, 2.60, 2.70) \\ (2.20, 2.30, 2.40) & (0.42, 0.43, 0.45) & (0.37, 0.38, 0.40) & (1.00, 1.00, 1.00) \end{pmatrix}$$

$$\tilde{B}^{(5)} = (\tilde{b}_{ij}^{(5)})_{4\times 4} = \begin{pmatrix} (1.00, 1.00, 1.00) & (1.60, 1.70, 1.80) & (1.00, 1.10, 1.30) & (2.00, 2.10, 2.20) \\ (0.50, 0.60, 0.63) & (1.00, 1.00, 1.00) & (0.60, 0.63, 0.67) & (1.70, 1.80, 1.90) \\ (0.77, 0.90, 1.00) & (1.50, 1.60, 1.70) & (1.00, 1.00, 1.00) & (1.80, 1.90, 2.00) \\ (0.45, 0.48, 0.50) & (0.53, 0.56, 0.60) & (0.50, 0.53, 0.56) & (1.00, 1.00, 1.00) \end{pmatrix}$$

$$\tilde{\Pi}^{(1)} = \left(\tilde{u}_{ij}^{(1)}(x)\right)_{5\times 4} = \begin{pmatrix} (0.3, 0.5, 0.6) & (0.4, 0.5, 0.7) & (0.5, 0.7, 0.8) & (0.4, 0.6, 0.7) \\ (0.2, 0.3, 0.5) & (0.3, 0.4, 0.5) & (0.4, 0.5, 0.7) & (0.3, 0.5, 0.6) \\ (0.2, 0.3, 0.4) & (0.4, 0.5, 0.6) & (0.4, 0.5, 0.6) & (0.3, 0.4, 0.5) \\ (0.4, 0.6, 0.7) & (0.4, 0.5, 0.6) & (0.3, 0.4, 0.6) & (0.5, 0.6, 0.7) \\ (0.6, 0.7, 0.8) & (0.5, 0.6, 0.7) & (0.3, 0.5, 0.6) & (0.6, 0.7, 0.8) \end{pmatrix} \quad (13)$$

$$\tilde{\Pi}^{(2)} = \left(\tilde{u}_{ij}^{(2)}(x)\right)_{5\times 4} = \begin{pmatrix} (0.4, 0.5, 0.6) & (0.5, 0.6, 0.7) & (0.6, 0.7, 0.8) & (0.4, 0.5, 0.6) \\ (0.3, 0.4, 0.5) & (0.4, 0.5, 0.6) & (0.3, 0.5, 0.7) & (0.4, 0.6, 0.7) \\ (0.3, 0.5, 0.6) & (0.4, 0.6, 0.7) & (0.4, 0.6, 0.7) & (0.3, 0.4, 0.6) \\ (0.6, 0.7, 0.8) & (0.3, 0.4, 0.6) & (0.3, 0.4, 0.5) & (0.5, 0.7, 0.8) \\ (0.6, 0.7, 0.8) & (0.6, 0.7, 0.8) & (0.3, 0.4, 0.5) & (0.7, 0.8, 0.9) \end{pmatrix} \quad (14)$$

$$\tilde{\Pi}^{(3)} = \left(\tilde{\mu}_{ij}^{(3)}(x)\right)_{5\times 4} = \begin{pmatrix} (0.4, 0.5, 0.6) & (0.5, 0.7, 0.8) & (0.5, 0.6, 0.7) & (0.5, 0.6, 0.7) \\ (0.3, 0.4, 0.5) & (0.4, 0.6, 0.7) & (0.4, 0.5, 0.6) & (0.4, 0.6, 0.7) \\ (0.2, 0.3, 0.5) & (0.6, 0.7, 0.8) & (0.4, 0.6, 0.7) & (0.3, 0.5, 0.6) \\ (0.6, 0.8, 0.9) & (0.3, 0.4, 0.5) & (0.3, 0.4, 0.5) & (0.6, 0.7, 0.8) \\ (0.7, 0.8, 0.9) & (0.6, 0.8, 0.9) & (0.3, 0.5, 0.7) & (0.6, 0.8, 0.9) \end{pmatrix} \quad (15)$$

$$\tilde{\Pi}^{(4)} = \left(\tilde{\mu}_{ij}^{(4)}(x)\right)_{5\times 4} = \begin{pmatrix} (0.4, 0.6, 0.7) & (0.6, 0.7, 0.8) & (0.6, 0.7, 0.9) & (0.6, 0.7, 0.8) \\ (0.4, 0.6, 0.8) & (0.4, 0.5, 0.7) & (0.3, 0.4, 0.5) & (0.3, 0.5, 0.7) \\ (0.3, 0.5, 0.7) & (0.7, 0.8, 0.9) & (0.5, 0.6, 0.7) & (0.2, 0.4, 0.5) \\ (0.7, 0.8, 0.9) & (0.3, 0.5, 0.6) & (0.2, 0.3, 0.4) & (0.7, 0.8, 0.9) \\ (0.7, 0.8, 0.9) & (0.6, 0.7, 0.9) & (0.3, 0.5, 0.6) & (0.7, 0.8, 0.9) \end{pmatrix} \quad (16)$$

$$\tilde{\Pi}^{(5)} = \left(\tilde{\mu}_{ij}^{(5)}(x)\right)_{5\times 4} = \begin{pmatrix} (0.3, 0.4, 0.5) & (0.5, 0.6, 0.7) & (0.6, 0.7, 0.8) & (0.5, 0.6, 0.7) \\ (0.2, 0.3, 0.4) & (0.2, 0.3, 0.4) & (0.2, 0.4, 0.5) & (0.3, 0.4, 0.5) \\ (0.2, 0.3, 0.4) & (0.5, 0.7, 0.9) & (0.6, 0.7, 0.8) & (0.2, 0.3, 0.4) \\ (0.4, 0.6, 0.8) & (0.5, 0.6, 0.7) & (0.6, 0.7, 0.8) & (0.4, 0.6, 0.7) \\ (0.7, 0.8, 0.9) & (0.2, 0.4, 0.6) & (0.2, 0.3, 0.4) & (0.3, 0.4, 0.5) \end{pmatrix} \quad (17)$$

First, determine the weight of criterias.

The optimization model to determine weight of criterias is as follows;

$$E = \min\left(\sum_{j=1}^{4} w_{J_1}(el_j^{(1)} + em_j^{(1)} + eu_j^{(1)}) + \sum_{i,j=1}^{4}\sum_{d=2}^{5} w_{J_d}(el_{ij}^{(d)} + em_{ij}^{(d)} + eu_{ij}^{(d)})\right)$$

s. t.

$$w_j \geq ul_j^{(1)} - el_j^{(1)}, \; j = \overline{1, 4}$$
$$w_j \leq uu_j^{(1)} + eu_j^{(1)}, \; j = \overline{1, 4}$$
$$el_j^{(1)}, eu_j^{(1)} \geq 0, \; j = \overline{1, 4}$$
$$0.5(w_i - w_j + 1) \geq pl_{ij}^{(d)} - el_{ij}^{(d)}, \; i, j = \overline{1, 4}, d = 2, 3$$
$$0.5(w_i - w_j + 1) \leq pu_{ij}^{(d)} + eu_{ij}^{(d)}, \; i, j = \overline{1, 4}, d = 2, 3$$

$$w_j \geq ul_j^{(1)} - el_j^{(1)}, \; j = \overline{1, 4}$$
$$\frac{w_i}{w_j} \leq bu_{ij}^{(d)} + eu_{ij}^{(d)}, \; i, j = \overline{1, 4}, d = 4, 5$$
$$el_{ij}^{(d)}, eu_{ij}^{(d)} \geq 0, \; i, j = \overline{1, 4}, d = \overline{2, 5}$$
$$\sum_{j=1}^{4} w_j = 1, w_j \geq 0, \; j = \overline{1, 4}$$

Where

$$em_j^{(1)} = \left(w_j - um_j^{(1)}\right)^2, \; j = \overline{1,4}, \quad em_{ij}^{(d)} = \left(0.5(w_i - w_j + 1) - pm_{ij}^{(d)}\right)^2, \; i, j = \overline{1, 4}, d = 2, 3,$$

$$em_{ij}^{(d)} = \left(\frac{w_i}{w_j} - bm_{ij}^{(d)}\right)^2, \; i, j = \overline{1, 4}, d = 4, 5$$

Above model is solved by using mathematical problem solving tool MATHEMATICA, then semi-optimal errors are as follows;

$eu_2^{(1)*} = 0.06$, $eu_{12}^{(2)*} = 0.12$, $eu_{13}^{(2)*} = 0.03$, $el_{14}^{(2)*} = 0.02$, $el_{23}^{(2)*} = 0.04$, $el_{24}^{(2)*} = 0.28$, $el_{34}^{(2)*} = 0.09$,

$eu_{12}^{(3)*} = 0.17$, $eu_{13}^{(3)*} = 0.03$, $eu_{14}^{(3)*} = 0.23$, $el_{23}^{(3)*} = 0.09$, $eu_{24}^{(3)*} = 0.02$, $eu_{34}^{(3)*} = 0.16$, $eu_{12}^{(4)*} = 0.32$,

$eu_{13}^{(4)*} = 0.21$, $eu_{14}^{(4)*} = 0.41$, $eu_{23}^{(4)*} = 0.15$, $el_{24}^{(4)*} = 1.04$, $el_{34}^{(4)*} = 0.99$, $el_{12}^{(5)*} = 0.87$, $el_{13}^{(5)*} = 0.43$,

$el_{14}^{(5)*} = 1.15$, $eu_{23}^{(5)*} = 0.10$, $el_{24}^{(5)*} = 0.54$, $el_{34}^{(5)*} = 0.29$.

The other errors are zero. Therefore, according to step 3, requirement information of decision makers are changed as follows;

$\tilde{u}_2^{(1)'} = (0.05, 0.15, 0.26)$, $\tilde{p}_{12}^{(2)'} = (0.20, 0.30, 0.47)$, $\tilde{p}_{13}^{(2)'} = (0.25, 0.30, 0.43)$, $\tilde{p}_{14}^{(2)'} = (0.48, 0.55, 0.60)$,

$\tilde{p}_{23}^{(2)'} = (0.46, 0.65, 0.75)$, $\tilde{p}_{24}^{(2)'} = (0.52, 0.85, 0.95)$, $\tilde{p}_{34}^{(2)'} = (0.56, 0.70, 0.80)$, $\tilde{p}_{12}^{(3)'} = (0.25, 0.25, 0.47)$,

$\tilde{p}_{13}^{(3)'} = (0.30, 0.35, 0.43)$, $\tilde{p}_{14}^{(3)'} = (0.15, 0.20, 0.48)$, $\tilde{p}_{23}^{(3)'} = (0.46, 0.60, 0.65)$, $\tilde{p}_{24}^{(3)'} = (0.40, 0.45, 0.52)$,

$\tilde{p}_{34}^{(3)'} = (0.25, 0.30, 0.56)$, $\tilde{b}_{12}^{(4)'} = (0.38, 0.40, 0.74)$, $\tilde{b}_{13}^{(4)'} = (0.33, 0.34, 0.57)$, $\tilde{b}_{14}^{(4)'} = (0.42, 0.43, 0.86)$,

$\tilde{b}_{23}^{(4)'} = (0.55, 0.59, 0.77)$, $\tilde{b}_{24}^{(4)'} = (1.16, 2.30, 2.40)$, $\tilde{b}_{34}^{(4)'} = (1.51, 2.60, 2.70)$, $\tilde{b}_{12}^{(5)'} = (0.73, 1.70, 1.80)$,

$\tilde{b}_{13}^{(5)'} = (0.57, 1.10, 1.30)$, $\tilde{b}_{14}^{(5)'} = (0.85, 2.10, 2.20)$, $\tilde{b}_{23}^{(5)'} = (0.60, 0.63, 0.77)$, $\tilde{b}_{24}^{(5)'} = (1.16, 1.80, 1.90)$,

$\tilde{b}_{34}^{(5)'} = (1.51, 1.90, 2.00)$.

According to step 4, we make the optimal model again, and get semi-optimal error again. After two playing these process, all errors are zero. At this time, semi-optimal weight vector $W^* = (0.189, 0.257, 0.333, 0.221)$ is just the optimal weight vector of this problem.

To evaluate the effectiveness of supported algorithm, we compared this weight vector with $W = (0.205, 0.259, 0.313, 0.223)$ which is gotten by precedent method[9].

The sum of errors that weight by precedent method and real weight is 1.226, the sum of errors that weight by supported method and real weight is 1.077. So supported determining method of weights reduces errors to 87.8%.

At this time, we evaluate the effectiveness of determining method of optimal alternative.

At first, we use the precedent method. If we use only infimum, then the result is $a_1 \succ a_5 \succ a_3 \succ a_2 \succ a_4$, if use only median, thent it is $a_3 \succ a_4 \succ a_1 \succ a_5 \succ a_2$, if we use only supremum, then it is $a_5 \succ a_4 \succ a_2 \succ a_1 \succ a_3$. That is, results are different.

Then we get the optimal alternative of this problem by using supported method.

Accorind to step 1, every decision maker calculates the cooperative degree and conflicting degree of criterias' pair, the result is as table 1.

Table 1. Cooperative degree and conflicting degree of criterias' pair

| cp, cf criterias' pair | $J_1$ | | $J_2$ | | $J_3$ | | $J_4$ | | $J_5$ | |
|---|---|---|---|---|---|---|---|---|---|---|
| | cp | cf | cp | cf | cp | cf | cp | cf | cp | cf |
| $(c_1, c_2)$ | 0.833 | 0.067 | 0.500 | 0.400 | 0.478 | 0.522 | 0.188 | 0.750 | 0.348 | 0.522 |
| $(c_1, c_3)$ | 0.294 | 0.706 | 0.188 | 0.812 | 0.132 | 0.763 | 0.139 | 0.806 | 0.208 | 0.604 |
| $(c_1, c_4)$ | 1.000 | 0.000 | 0.833 | 0.139 | 1.000 | 0.000 | 0.947 | 0.053 | 0.571 | 0.286 |
| $(c_2, c_3)$ | 0.500 | 0.450 | 0.567 | 0.333 | 0.786 | 0.143 | 0.889 | 0.111 | 0.905 | 0.048 |
| $(c_2, c_4)$ | 0.818 | 0.091 | 0.529 | 0.471 | 0.531 | 0.469 | 0.263 | 0.684 | 0.500 | 0.472 |
| $(c_3, c_4)$ | 0.385 | 0.615 | 0.056 | 0.917 | 0.167 | 0.750 | 0.310 | 0.643 | 0.579 | 0.237 |

We divide overall criterias' set by using supported dividing algorithm in this paper.

Table 2. Dividing result of criterias' set

| J subset | $J_1$ | $J_2$ | $J_3$ | $J_4$ | $J_5$ |
|---|---|---|---|---|---|
| $S_1$ | $c_1, c_4$ | $c_1, c_4$ | $c_1, c_4$ | $c_1, c_4$ | $c_1, c_4$ |
| $S_2$ | $c_2, c_3$ | $c_2$ | $c_2$ | $c_2, c_3$ | $c_2$ |
| $S_3$ | | $c_3$ | $c_3$ | | $c_3$ |

The final degree of satisfaction is as table 3.

Table 3. the final degree of satisfaction

| J A | $J_1$ | $J_2$ | $J_3$ | $J_4$ | $J_5$ |
|---|---|---|---|---|---|
| $a_1$ | (0.278,0.356,0.448) | (0.293,0.359,0.426) | (0.315,0.407,0.474) | (0.354,0.413,0.505) | (0.328,0.387,0.446) |
| $a_2$ | (0.181,0.270,0.356) | (0.248,0.337,0.403) | (0.248,0.362,0.429) | (0.203,0.262,0.346) | (0.118,0.210,0.269) |
| $a_3$ | (0.207,0.274,0.340) | (0.226,0.337,0.426) | (0.258,0.347,0.433) | (0.346,0.405,0.464) | (0.328,0.413,0.498) |
| $a_4$ | (0.289,0.375,0.441) | (0.301,0.390,0.482) | (0.323,0.409,0.475) | (0.144,0.228,0.287) | (0.328,0.387,0.446) |
| $a_5$ | (0.375,0.441,0.508) | (0.422,0.489,0.556) | (0.419,0.534,0.600) | (0.254,0.346,0.431) | (0.118,0.203,0.287) |

The intergrated final degree of satisfaction is as follows:

$$\mu(a_1) = (0.306, 0.383, 0.459)$$

$$\mu(a_2) = (0.206, 0.295, 0.367)$$

$$\mu(a_3) = (0.268, 0.351, 0.429)$$

$$\mu(a_4) = (0.276, 0.358, 0.428)$$

$$\mu(a_5) = (0.333, 0.417, 0.490)$$

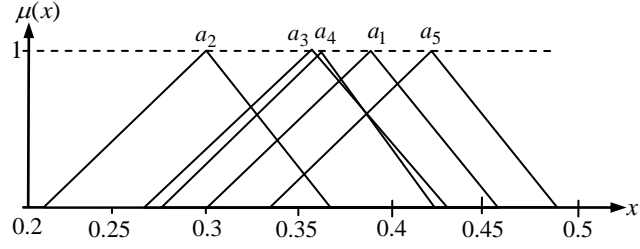

The integrated final degrees of satisfactrion are arranged by using fuzzy arranging function. Here let be $p=1$, $q=1/2$, $a_2 = a_3$, $P_L = 1$, $P_R = 1$. Then

$$F_{1,1/2}(\mu(a_1)) = \frac{0.306 + 2 \times 0.383 + 0.459}{4} = 0.383$$

$$F_{1,1/2}(\mu(a_2)) = \frac{0.206 + 2 \times 0.295 + 0.367}{4} = 0.291$$

$$F_{1,1/2}(\mu(a_3)) = \frac{0.268 + 2 \times 0.351 + 0.429}{4} = 0.350$$

$$F_{1,1/2}(\mu(a_4)) = \frac{0.276 + 2 \times 0.358 + 0.428}{4} = 0.355$$

$$F_{1,1/2}(\mu(a_5)) = \frac{0.333 + 2 \times 0.417 + 0.490}{4} = 0.414$$

$$F_{1,\frac{1}{2}}(\mu(a_5)) > F_{1,\frac{1}{2}}(\mu(a_1)) > F_{1,\frac{1}{2}}(\mu(a_4)) > F_{1,\frac{1}{2}}(\mu(a_3)) > F_{1,\frac{1}{2}}(\mu(a_2))$$

Therefore

$$\mu(a_5) \succ \mu(a_1) \succ \mu(a_4) \approx \mu(a_3) \succ \mu(a_2)$$

In other words, the optimal alternative is $a_5$, the fifth alternative is the best one.

The precedent method cann't consider various conditions, so its result is not stabilitied. But the suggested method in this paper is so considering all possible conditions that its result is stabilitied.

### 5. CONCLUSION

In this paper, we supported a model to determine the weight of criterias based on considering infimum and supremum of information in case that the requirement information of decision makers were given in various forms, and suggested a dividing algorithm of criterias' set to determine the optimal alternative. Also we defined a new final evaluation criterion and evaluated the effectiveness of suggested method through solving actual problem.